\documentclass[journal,twoside]{IEEEtran}
\usepackage{cite}
\usepackage{amsmath,amssymb,amsfonts}
\usepackage{graphicx}
\usepackage{booktabs}
\usepackage{multirow}
\usepackage{array}
\usepackage{tabularx}
\usepackage{xcolor}

\newcommand{\F}{\mathbf{F}}
\newcommand{\R}{\mathbb{R}}
\newcommand{\y}{\mathbf{y}}
\newcommand{\p}{\mathbf{p}}
\newcommand{\Pose}{\mathbf{P}}

\newcommand{\abstracttext}{%
A camera becomes an absolute pose sensor when its images can be relocalized against a prebuilt map of the scene. On mobile robots and augmented-reality devices, that map must be small and quick to build. Scene coordinate regression (SCR) stores the scene in a compact network that predicts a 3D coordinate for every pixel, and mixture-of-experts (MoE) variants extend its capacity to larger outdoor scenes. In existing MoE-based SCR, however, the router assigns each pixel to an expert from the same local descriptor that the experts regress from, so repeated facades and windows at different locations receive similar routing inputs. We present DAC-MACE, which conditions soft expert routing on recalibrated and contextualized features. A residual context adapter (RCA) reweights the channels and locations of frozen encoder features, a global context encoder (GCE) mixes them with multi-scale self-attention, and an image-level pooled descriptor is added to the router input. A dual-threshold RANSAC--Levenberg--Marquardt estimator recovers the pose. On Cambridge Landmarks, DAC-MACE attains a mean per-scene median translation error of 13.0\,cm over three seeds, the lowest among the RGB-only SCR methods compared (GLACE 13.6\,cm, MACE 14.4\,cm), and reduces the error on the largest scene from 24 to 19\,cm relative to MACE. In a cumulative study, the MoE head, GCE, RCA, and the dual-threshold estimator lower the mean error from 17.0\,cm to 14.4, 14.0, 13.6, and 13.0\,cm. On 7-Scenes, adding GCE to a single ACE head raises Recall@(5\,cm,\,5$^\circ$) from 86.56\% to 87.60\%.}

\newcommand{\indextermstext}{%
Camera relocalization, image sensors, mixture of experts, robust pose estimation, scene coordinate regression, self-attention, visual localization.}

\begin{document}

\title{DAC-MACE: Context-Conditioned Expert Routing for Compact Scene Coordinate Regression in Camera Relocalization}

\author{Dingying~Fan$^{1,\dagger}$, Mingkai~Liu$^{2,\dagger}$, Haohua~Que$^{1,3,\dagger}$, Haojia~Gao$^{4}$, Xianliang~Huang$^{5}$, Qian~Zhang$^{1,3}$, Xueqian~Wang$^{4}$, and~Fei~Qiao$^{1,*}$
\thanks{This work has been submitted to the IEEE for possible publication. Copyright may be transferred without notice, after which this version may no longer be accessible.}
\thanks{This work was supported by the National Natural Science Foundation of China (U25A20489, 62334006), the Beijing Natural Science Foundation (L253009), the National Science and Technology Major Project Fund of China (2025ZD0215600), the National Key Technologies R\&D Program of China (2025YFF1500600), and the Fundamental Research Program of Shanxi Province (202403021212166, 202303021211153). $^{\dagger}$Equal contribution. $^{*}$Corresponding author: Fei Qiao (e-mail: qiaofei@tsinghua.edu.cn).}
\thanks{$^{1}$Sense Lab, Dept. of Electronic Engineering, Tsinghua University, Beijing 100084, China. $^{2}$School of Software and Microelectronics, Peking University, Beijing 102600, China. $^{3}$Infinity Robotics, Beijing, China. $^{4}$Tsinghua Shenzhen International Graduate School, Tsinghua University, Shenzhen 518055, China. $^{5}$PICO, ByteDance Inc., Beijing, China.}}

\maketitle
\pagestyle{plain}
\thispagestyle{plain}

\begin{abstract}
\abstracttext
\end{abstract}


% =====================================================================
\section{Introduction}
\label{sec:intro}

\IEEEPARstart{A}{} calibrated camera is the most widely deployed exteroceptive sensor on mobile robots, vehicles, and head-mounted devices. Visual relocalization turns it into an absolute pose sensor: given a single query image, it estimates the six-degree-of-freedom (6-DoF) camera pose with respect to a previously mapped scene~\cite{kendall2015posenet,sattler2018benchmarking}. Such absolute fixes bound the drift of visual--inertial odometry and substitute for satellite positioning where it is unavailable or degraded, for example indoors and between tall buildings. Accuracy is not the only requirement in these deployments. The scene representation must fit into device memory, it must be rebuilt quickly when a site is remapped, and the per-query computation must fit the platform budget.

Structure-based pipelines match local features against a structure-from-motion (SfM) point cloud~\cite{sattler2012improving,sarlin2019hloc} and are accurate, but their maps occupy hundreds of megabytes (Table~\ref{tab:cambridge}). Scene coordinate regression (SCR) stores the scene implicitly in a network that predicts a 3D scene coordinate for every pixel; the pose then follows from the 2D--3D correspondences with RANSAC and a perspective-$n$-point (PnP) solver~\cite{shotton2013scene,brachmann2017dsac,brachmann2022dsacstar}. ACE~\cite{brachmann2023ace} reduced SCR mapping to minutes and the map to 4\,MB by training only a small regression head on the features of a frozen, scene-agnostic encoder. The capacity of a single head is shared by the entire scene, however, and on Cambridge Landmarks~\cite{kendall2015posenet} the median translation error of ACE grows from 5\,cm on the smallest scene to 43\,cm on the largest.

Partitioning the scene and assigning a regressor to each part raises capacity. ESAC~\cite{brachmann2019esac} and the Poker ensemble of ACE~\cite{brachmann2023ace} rely on fixed partitions, while MACE~\cite{liu2025mace} learns a router that weights ACE experts pretrained on spatial clusters. In all three, the routing decision is computed from the same local encoder descriptor the experts regress from. Each descriptor covers an $81{\times}81$-pixel receptive field, so repeated facades, windows, and weakly textured surfaces at different locations produce similar descriptors and hence similar routing inputs~\cite{torii2015place}. We refer to this as \emph{perceptual routing ambiguity}. It differs from \emph{expert-load collapse}, in which assignments concentrate on a few experts~\cite{shazeer2017moe,fedus2022switch}. Load balancing regularizes the marginal usage of the experts, but it does not tell the router which expert is geometrically appropriate for a given pixel.

This paper studies whether conditioning the router on information beyond the local descriptor improves MoE-based SCR. We propose DAC-MACE (Dynamic Attention-driven Context MACE, Fig.~\ref{fig:overview}). Two lightweight trainable modules are placed between the frozen ACE encoder and the experts: a residual context adapter (RCA) that rescales the channels and locations of the encoder features, and a global context encoder (GCE) that mixes features with multi-scale self-attention. An image-level pooled descriptor is further added to the router input, and the pose is recovered with a dual-threshold RANSAC--LM estimator. The design keeps the buffer-based training of ACE, which bounds the mapping time; Section~\ref{sec:protocol} states precisely what context RCA and GCE observe under this training scheme.

The contributions of this paper are as follows.
\begin{itemize}
\item A context-conditioned soft MoE for SCR that keeps the encoder frozen and the ACE training pipeline intact, together with an explicit account of the context each module observes during training and at test time.
\item A dual-threshold RANSAC--LM pose stage with an interpretable selection rule, evaluated on fixed trained models and analyzed through per-scene branch statistics.
\item An evaluation on Cambridge Landmarks and 7-Scenes covering three-seed accuracy, a component-wise analysis, qualitative trajectory and feature-response diagnostics, expert usage, and model complexity. Among the RGB-only SCR methods compared, DAC-MACE has the lowest mean of per-scene median translation errors, every added component lowers that mean, and the largest gain over MACE is on the largest scene (Great Court, 24 to 19\,cm).
\end{itemize}
% =====================================================================
\section{Related Work}
\label{sec:related}

\subsection{Structure-Based Localization and Retrieval}
Feature-matching pipelines establish 2D--3D correspondences between query keypoints and an SfM model~\cite{sattler2012improving,sarlin2019hloc,sarlin2020superglue}, often after image retrieval narrows the search~\cite{arandjelovic2016netvlad,torii2015place}. InLoc~\cite{taira2018inloc} extends this to indoor scenes with dense matching, CAMNet~\cite{ding2019camnet} combines coarse-to-fine retrieval with relative pose estimation, and PixLoc~\cite{sarlin2021pixloc} aligns learned feature maps directly. These methods are accurate but store descriptors for every 3D point; compression~\cite{camposeco2019hybrid} or descriptor-free matching~\cite{zhou2022gomatch} reduces the map at a large cost in accuracy (Table~\ref{tab:cambridge}). Absolute pose regressors~\cite{kendall2015posenet,kendall2017geometric,shavit2021mstransformer} are compact but considerably less accurate.

\subsection{Scene Coordinate Regression}
The DSAC family~\cite{brachmann2017dsac,brachmann2018learning,brachmann2022dsacstar} learns per-pixel scene coordinates end to end with differentiable RANSAC. HSCNet~\cite{li2020hscnet} and HSCNet++~\cite{wang2024hscnetpp} predict coordinates hierarchically, first classifying the scene region and then regressing within it, and HSCNet++ adds transformer-based conditioning across levels. SRC~\cite{dong2022src} replaces regression with few-shot region classification, and SANet~\cite{yang2019sanet} regresses coordinates in a scene-agnostic manner. ACE~\cite{brachmann2023ace} trains a scene-specific head on a buffer of frozen encoder features and maps a scene in about five minutes. GLACE~\cite{wang2024glace} concatenates a pretrained image-retrieval descriptor to every local feature and predicts coordinates through a position decoder over clustered camera centers, which lets a single head scale to larger scenes. R-SCoRe~\cite{jiang2025rscore} revisits the training of SCR for large-scale benchmarks such as Aachen Day--Night~\cite{sattler2018benchmarking}. The image-level residual of our router is close in spirit to the global descriptor of GLACE; it differs in that it pools the scene-adapted features rather than a separately pretrained retrieval feature, and in that it conditions the router rather than the regressor.

\subsection{Partitioned and Mixture-of-Experts Regression}
ESAC~\cite{brachmann2019esac} combines a gating network with expert regressors and samples hypotheses across experts. Poker~\cite{brachmann2023ace} trains independent ACE heads on spatial clusters and selects the pose with the most inliers. MACE~\cite{liu2025mace}, the direct predecessor of this work, pretrains ACE experts on clusters of mapping-camera positions, routes features with a Gumbel-softmax gate~\cite{jang2017gumbel}, and balances the experts with an auxiliary-loss-free bias update~\cite{wang2024auxfree}. DAC-MACE retains the experts, the position decoder, and the bias update of MACE. It changes the router input (RCA, GCE, and the pooled residual), adds a standard-deviation penalty on mean expert usage, evaluates all experts with soft weights at test time, and replaces the pose stage. In the broader MoE literature, load imbalance is countered with auxiliary losses~\cite{shazeer2017moe,fedus2022switch,riquelme2021vmoe}; routing ambiguity caused by uninformative router inputs has received less attention in SCR.

\subsection{Robust Pose Estimation}
RANSAC~\cite{fischler1981ransac} remains the standard for PnP with outliers. LO-RANSAC~\cite{chum2003lo} interleaves local optimization with hypothesis sampling, MAGSAC++~\cite{barath2020magsacpp} marginalizes over the inlier threshold, and graduated non-convexity~\cite{yang2020gnc} solves robust estimation without sampling. DSAC*~\cite{brachmann2022dsacstar} and ACE score hypotheses with a soft inlier count and refine the best one iteratively. Our dual-threshold stage is a lightweight heuristic in this space that targets the varying coordinate accuracy of SCR predictions.

% =====================================================================
\section{Method}
\label{sec:method}

\begin{figure*}[t]
\centering
\includegraphics[width=\textwidth]{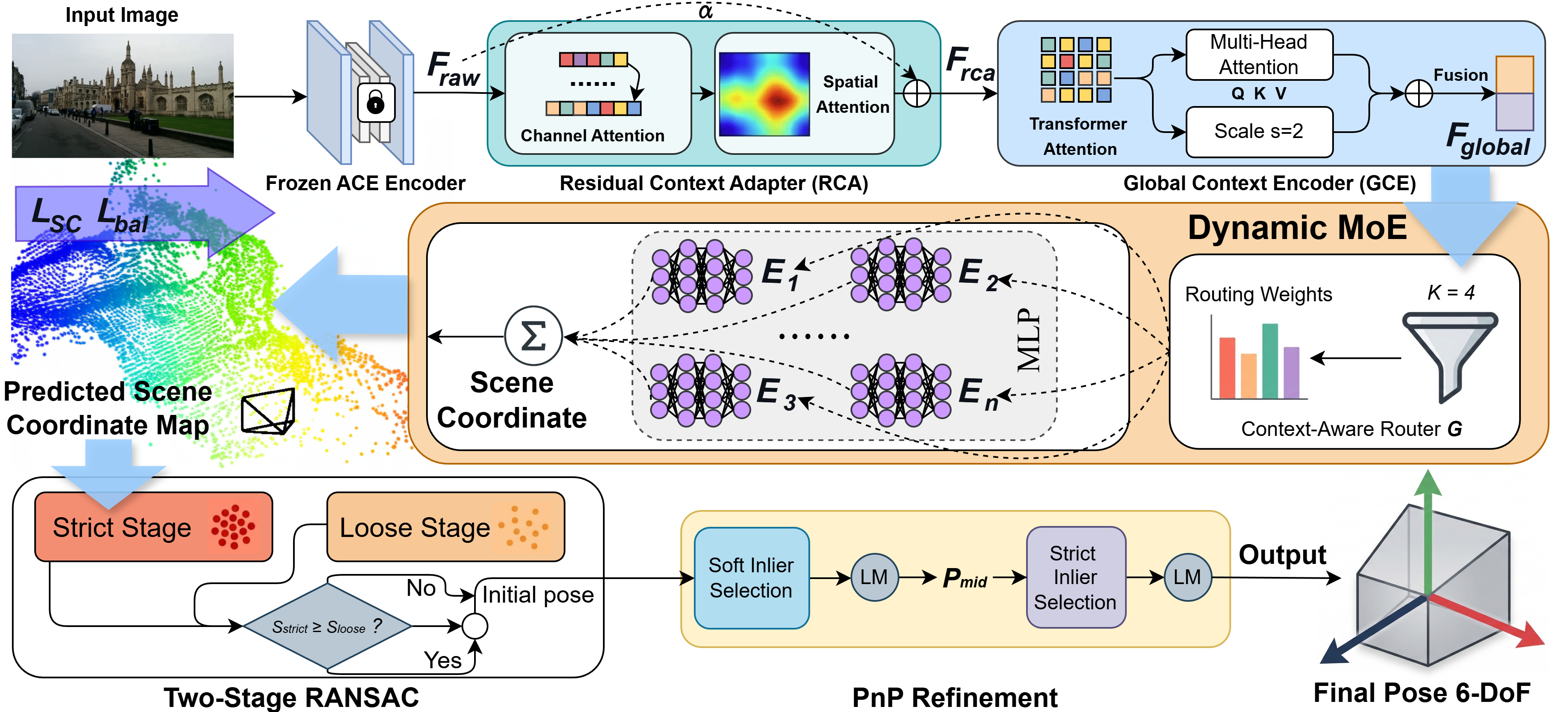}
\caption{DAC-MACE pipeline. The frozen ACE encoder produces $\F_{\mathrm{raw}}$. RCA recalibrates channels and locations ($\F_{\mathrm{rca}}$), and GCE adds multi-scale self-attention context ($\F_{\mathrm{g}}$). The router receives $\F_{\mathrm{g}}$ plus an image-level pooled residual and predicts per-pixel soft weights over $K$ position-aware expert heads $E_1,\dots,E_K$, whose fused scene coordinates are passed to the dual-threshold RANSAC--LM pose stage. Training uses the reprojection loss $\mathcal{L}_{\mathrm{sc}}$ and the load-balancing term $\mathcal{L}_{\mathrm{bal}}$.}
\label{fig:overview}
\end{figure*}

\subsection{Problem Formulation and Overview}

Given a grayscale query image $\mathbf{I}\in\R^{H\times W}$ from a calibrated camera with intrinsics $\mathbf{K}_{\mathrm{cam}}$, we estimate its pose $\Pose=(\mathbf{R},\mathbf{t})\in SE(3)$. SCR learns a mapping $f$ from each pixel $\p$ to a scene coordinate $\y_{\p}\in\R^3$ and obtains the pose robustly from the correspondences,
\begin{equation}
\y_{\p}=f(\mathbf{I},\p),\qquad \Pose=\mathrm{PnP}\big(\{(\p,\y_{\p})\},\mathbf{K}_{\mathrm{cam}}\big).
\label{eq:scr}
\end{equation}
In large outdoor scenes, repeated structures make local descriptors ambiguous and increase the risk of coordinate outliers. DAC-MACE therefore consists of a frozen encoder, RCA, GCE, a context-conditioned router with $K$ experts, and a dual-threshold pose stage (Fig.~\ref{fig:overview}). Table~\ref{tab:notation} summarizes the notation; each symbol has a single meaning throughout the paper.

\begin{table}[t]
\centering
\caption{Notation}
\label{tab:notation}
\footnotesize
\begin{tabularx}{\columnwidth}{@{}lX@{}}
\toprule
Symbol & Meaning \\
\midrule
$C$, $H'{\times}W'$ & feature channels (512), feature-map size ($H/8{\times}W/8$) \\
$\F_{\mathrm{raw}},\F_{\mathrm{rca}},\F_{\mathrm{g}}$ & encoder, RCA, and GCE features \\
$K$, $k$ & number of experts, expert index \\
$g_k(\p)$ & routing weight of expert $k$ at pixel $\p$ \\
$\gamma_{\mathrm{r}}$, $\boldsymbol{\gamma}_{r}$, $\boldsymbol{\gamma}_{\mathrm{f}}$ & RCA gate; GCE branch and fusion LayerScales \\
$r\in\{1,2\}$ & GCE pooling scale \\
$\eta$ & weight of the router's pooled residual \\
$M$, $\mathbf{c}_{k,m}$ & number of stored centers per expert, centers \\
$B$ & number of buffer features per training step \\
$\omega(t)$ & soft-clamp threshold of the reprojection loss \\
$\tau_{\mathrm{s}},\tau_{\mathrm{l}}$ & strict and loose RANSAC inlier thresholds (px) \\
$\rho(\Pose,\tau)$ & threshold-normalized support of hypothesis $\Pose$ \\
\bottomrule
\end{tabularx}
\end{table}

\subsection{Frozen Encoder}

We use the pretrained ACE encoder~\cite{brachmann2023ace}, a fully convolutional ResNet-style network with stride 8, and keep it frozen:
\begin{equation}
\F_{\mathrm{raw}}=\mathcal{E}(\mathbf{I})\in\R^{C\times H'\times W'},\qquad C=512.
\end{equation}
Each output vector has an $81{\times}81$-pixel receptive field in the input image. Freezing the encoder preserves its scene-agnostic representation and allows its features to be precomputed once into a training buffer, which is the basis of the short mapping time of ACE.

\subsection{Residual Context Adapter}
\label{sec:rca}

The encoder is trained once for all scenes, so its channels are not tuned to any particular one. RCA is a per-scene adapter that reweights them before routing. Following CBAM~\cite{woo2018cbam}, a channel gate and a spatial gate are applied in sequence,
\begin{align}
\mathbf{M}_{\mathrm{c}}&=\sigma\big(\phi(\mathrm{AvgPool}(\F_{\mathrm{raw}}))+\phi(\mathrm{MaxPool}(\F_{\mathrm{raw}}))\big)\in\R^{C},\\
\F_{\mathrm{c}}&=\mathbf{M}_{\mathrm{c}}\odot\F_{\mathrm{raw}},\\
\mathbf{M}_{\mathrm{s}}&=\sigma\big(\kappa_{7\times7}*[\mathrm{Avg}_{C}(\F_{\mathrm{c}});\mathrm{Max}_{C}(\F_{\mathrm{c}})]\big)\in\R^{H'\times W'},
\end{align}
where $\phi$ is a shared two-layer MLP with reduction ratio 16, $\mathrm{Avg}_C$ and $\mathrm{Max}_C$ pool over channels, $\kappa_{7\times7}$ is a $7{\times}7$ convolution, and $\sigma$ is the sigmoid. Because a randomly initialized multiplicative gate applied to frozen features can destabilize the early iterations, the output is blended with a learnable scalar,
\begin{equation}
\F_{\mathrm{rca}}=\F_{\mathrm{raw}}+\gamma_{\mathrm{r}}\big(\mathbf{M}_{\mathrm{s}}\odot\F_{\mathrm{c}}-\F_{\mathrm{raw}}\big),\qquad \gamma_{\mathrm{r}}^{(0)}=0.01,
\label{eq:rca}
\end{equation}
so that RCA starts close to the identity and keeps the randomly initialized adapter near the frozen feature distribution at the beginning of training. The channel gate emphasizes encoder channels that are informative for the mapped scene, and the spatial gate scales each feature vector according to its channel-pooled response, attenuating weak or uninformative responses before they reach the router.

\subsection{Global Context Encoder}
\label{sec:gce}

RCA does not enlarge the receptive field, so a feature on a repeated window is still described only by its $81{\times}81$-pixel neighborhood. GCE lets every feature attend to all other features of the image through content-based self-attention without positional encoding~\cite{vaswani2017attention,wang2018nonlocal}. For scale $r$, the input $\F_{\mathrm{rca}}$ is average-pooled by $r{\times}r$, flattened to $\mathbf{X}_r\in\R^{N_r\times C}$, and processed by a pre-normalized eight-head attention block,
\begin{equation}
\mathbf{A}_r=\mathbf{X}_r+\boldsymbol{\gamma}_r\odot\mathrm{MHA}\big(\mathrm{LN}(\mathbf{X}_r)\big),\qquad r\in\{1,2\},
\label{eq:gce_branch}
\end{equation}
where $\mathrm{MHA}$ computes, for each head,
\begin{equation}
\mathrm{Attention}(\mathbf{Q},\mathbf{K}_{\mathrm{a}},\mathbf{V})=\mathrm{softmax}\!\big(\mathbf{Q}\mathbf{K}_{\mathrm{a}}^{\top}/\sqrt{d}\big)\mathbf{V},
\label{eq:attention}
\end{equation}
with head dimension $d=C/8$ and $\mathbf{Q},\mathbf{K}_{\mathrm{a}},\mathbf{V}$ linear projections of $\mathrm{LN}(\mathbf{X}_r)$. The scale-2 output is bilinearly upsampled to $H'{\times}W'$, and the branches are fused as
\begin{equation}
\F_{\mathrm{g}}=\F_{\mathrm{rca}}+\boldsymbol{\gamma}_{\mathrm{f}}\odot\psi\big([\mathbf{A}_1;\mathrm{Up}(\mathbf{A}_2)]\big),
\label{eq:gce}
\end{equation}
where $\psi$ is a $1{\times}1$ convolution with batch normalization and ReLU. The convolution weights of $\psi$ are initialized to zero and all LayerScale vectors~\cite{touvron2021cait} to 0.01, so GCE initially passes $\F_{\mathrm{rca}}$ through unchanged. Attention dropout of 0.1 and stochastic depth~\cite{huang2016droppath} regularize the per-scene training. The scale-1 branch relates individual features, whereas the scale-2 branch attends over $2{\times}2$-pooled features, which are four times fewer and summarize larger regions; the two-scale construction follows ASpanFormer~\cite{chen2022aspanformer}.

\subsection{Context-Conditioned Router and Position-Aware Experts}
\label{sec:moe}

A single regression head has limited capacity in large scenes~\cite{liu2025mace}, while fixed, pre-partitioned ensembles such as Poker~\cite{brachmann2023ace} inherit the boundaries of the initial clustering. We therefore use a dynamic MoE with $K$ experts and a context-aware router.

\emph{Router.} The router receives the GCE features plus an image-level residual,
\begin{equation}
\F_{\mathrm{router}}=\F_{\mathrm{g}}+\eta\,\mathrm{GAP}(\F_{\mathrm{g}}),\qquad \eta^{(0)}=0,
\end{equation}
where $\mathrm{GAP}$ denotes global average pooling over the spatial dimensions, broadcast back to every location. A four-layer MLP with hidden widths 512, 512, and 256, layer normalization, ReLU, and dropout 0.05 predicts $K$ logits $\mathbf{z}(\p)$ per pixel, to which a per-expert bias $\mathbf{b}$ is added. Training uses the Gumbel-softmax relaxation~\cite{jang2017gumbel} with temperature 1, and inference uses the softmax, yielding weights $\mathbf{g}(\p)$ on the probability simplex. The resulting weights define a continuous partition of the scene.

\emph{Experts.} Each expert is an ACE regression head~\cite{brachmann2023ace} with the position decoder of GLACE and MACE~\cite{wang2024glace,liu2025mace}: it predicts softmax weights $\mathbf{a}_k(\p)$ over $M=50$ stored scene centers $\{\mathbf{c}_{k,m}\}$ and adds a coordinate residual $\mathbf{u}_k(\p)$,
\begin{equation}
\hat{\y}^{k}_{\p}=\mathbf{u}_k(\p)+\sum_{m=1}^{M}a_{k,m}(\p)\,\mathbf{c}_{k,m}.
\label{eq:expert}
\end{equation}
When an expert predicts homogeneous coordinates, the output is dehomogenized before fusion. Following MACE, expert $k$ is initialized from an ACE head trained on the $k$-th cluster of mapping-camera positions. All experts take $\F_{\mathrm{g}}$ as input, and the final coordinate is the routing-weighted sum
\begin{equation}
\y_{\p}=\sum_{k=1}^{K}g_k(\p)\,\hat{\y}^{k}_{\p}.
\label{eq:fusion}
\end{equation}
Because routing is soft, all $K$ experts are evaluated at every pixel; DAC-MACE thus increases capacity without sparse-compute savings, and its cost is analyzed in Section~\ref{sec:complexity}.

\emph{Number of experts.} We use $K{=}2$ for the two Cambridge scenes with the smallest mapped area, Shop Facade (875\,m$^2$) and Old Hospital (2{,}000\,m$^2$), and $K{=}4$ for King's College (5{,}600\,m$^2$), St Mary's Church (4{,}800\,m$^2$), and Great Court (8{,}000\,m$^2$); the areas are those given in~\cite{kendall2015posenet}.

\subsection{Training Objective}
\label{sec:objective}

Following ACE, valid predictions are supervised with the dynamic-tanh reprojection loss. For a pixel with reprojection error $e_{\p}$ at training step $t$ of $T$,
\begin{equation}
\mathcal{L}_{\mathrm{repro}}(e_{\p},t)=\omega(t)\tanh\!\big(e_{\p}/\omega(t)\big),
\label{eq:repro}
\end{equation}
where the soft-clamp threshold decreases from $\omega_{\max}=50$\,px to $\omega_{\min}=1$\,px as
\begin{equation}
\omega(t)=\omega_{\min}+\Big(1-\sqrt{1-(t/T)^2}\Big)(\omega_{\max}-\omega_{\min}).
\end{equation}
Invalid predictions (behind the camera, too far, or with a reprojection error above a hard clamp) receive an $\ell_1$ penalty $\mathcal{L}_{\mathrm{invalid}}$ toward a fixed-depth target in camera coordinates~\cite{brachmann2023ace,brachmann2022dsacstar}. A small penalty on the spread of mean expert usage is added,
\begin{equation}
\mathcal{L}=\underbrace{\mathcal{L}_{\mathrm{repro}}+\mathcal{L}_{\mathrm{invalid}}}_{\mathcal{L}_{\mathrm{sc}}}+\lambda\underbrace{\mathrm{std}_{k}\big(\bar g_k\big)}_{\mathcal{L}_{\mathrm{bal}}},\qquad \bar g_k=\frac{1}{B}\sum_{\p}g_k(\p),
\label{eq:loss}
\end{equation}
with $\lambda=0.01$. In addition, the router bias is updated without gradients after every step, as in the auxiliary-loss-free scheme of MACE~\cite{liu2025mace,wang2024auxfree}: an exponential moving average $\bar u_k$ of $\bar g_k$ (momentum 0.9) shifts $b_k\leftarrow b_k-0.1\,(\bar u_k-1/K)$. No routing-entropy term is optimized; routing entropy is computed only as a diagnostic (Section~\ref{sec:routing}).

\subsection{Training Protocol}
\label{sec:protocol}

We keep the two-step, buffer-based protocol of ACE~\cite{brachmann2023ace}.
\begin{enumerate}
\item \emph{Buffer fill.} The frozen encoder is run in evaluation mode without gradients over the augmented mapping images. For each image, 1{,}024 valid feature vectors are written, together with their pixel coordinates, inverse camera pose, and intrinsics, into a shuffled CPU buffer of 64\,M entries; features are stored in 16-bit precision. The buffer holds raw encoder features, i.e., before RCA and GCE.
\item \emph{Context-routing stage.} Each training step draws $B=163{,}840$ buffer entries, reshapes them into $B/512$ blocks of $16{\times}32$ features, and applies RCA, GCE, the router, and the experts. The experts are trained sequentially: for each $k$, expert $k$, the router, RCA, and GCE are updated for 16 epochs while the encoder and the other experts stay frozen.
\end{enumerate}
We use AdamW with weight decay $10^{-4}$ and no gradient clipping, a one-cycle learning-rate schedule with peak rates of $5{\times}10^{-3}$ for the active expert, RCA, and GCE and $5{\times}10^{-4}$ for the router, and mixed precision. Apart from $K$, all training and pose-estimation settings are shared across scenes; Table~\ref{tab:hyper} lists them.

\emph{What context the modules observe.} The $16{\times}32$ blocks in step 2 are not image regions: their 512 entries are drawn from different mapping images and locations. During training, therefore, the channel gate of RCA pools statistics over a random set of scene features; the spatial gate and the $2{\times}2$ pooling of the scale-2 GCE branch combine unrelated neighbors; and scale-1 attention, which is permutation-equivariant because it has no positional encoding, aggregates over features from many images. At test time, the same operators act on the coherent feature map of one image. GCE is therefore trained as a content-based aggregation over sets of scene features and applied within a single view at test time. This preserves the fast buffer-based training, and the results in Section~\ref{sec:exp} show that the resulting modules improve accuracy on average; a training scheme with image-coherent feature windows is left to future work (Section~\ref{sec:limitations}).

\begin{table*}[t]
\centering
\caption{Cambridge Landmarks: Median Translation (cm) / Rotation ($^\circ$) Errors. Avg. Is the Mean of the Five Per-Scene Medians, Computed From the Values Listed. Bold Marks the Best RGB-Only SCR Translation Error Per Column. Mapping Times Come From Different Implementations and Hardware and Are Indicative Only. $^{\dagger}$Excludes Pre-Clustering Time. $^{\ddagger}$Routing Stage, Excluding the ACE Pretraining of the Cluster Experts. $^{\S}$Expert Heads; for DAC-MACE, $K{=}4$ (5\,MB for $K{=}2$), With the Context Modules Listed in Table~\ref{tab:complexity}. The DAC-MACE Row Is the Mean Over Three Seeds.}
\label{tab:cambridge}
\footnotesize
\setlength{\tabcolsep}{2pt}
\begin{tabular*}{\textwidth}{@{\extracolsep{\fill}}ll c c c ccccc c@{}}
\toprule
& Method & \begin{tabular}[c]{@{}c@{}}Mesh/\\Depth\end{tabular} & \begin{tabular}[c]{@{}c@{}}Mapping\\time\end{tabular} & \begin{tabular}[c]{@{}c@{}}Map\\size\end{tabular} & \begin{tabular}[c]{@{}c@{}}Great\\Court\end{tabular} & \begin{tabular}[c]{@{}c@{}}King's\\College\end{tabular} & \begin{tabular}[c]{@{}c@{}}Old\\Hospital\end{tabular} & \begin{tabular}[c]{@{}c@{}}Shop\\Facade\end{tabular} & \begin{tabular}[c]{@{}c@{}}St Mary's\\Church\end{tabular} & \begin{tabular}[c]{@{}c@{}}Avg.\\(cm/$^\circ$)\end{tabular} \\
\midrule
\multirow{5}{*}{\rotatebox{90}{FM}}
& AS (SIFT)~\cite{sattler2012improving} & No & -- & ${\sim}$200\,MB & 24/0.1 & 13/0.2 & 20/0.4 & 4/0.2 & 8/0.3 & 13.8/0.24 \\
& hLoc (SP+SG)~\cite{sarlin2019hloc,sarlin2020superglue} & No & -- & ${\sim}$800\,MB & 16/0.1 & 12/0.2 & 15/0.3 & 4/0.2 & 7/0.2 & 10.8/0.20 \\
& PixLoc~\cite{sarlin2021pixloc} & No & ${\sim}$35\,min & ${\sim}$600\,MB & 30/0.1 & 14/0.2 & 16/0.3 & 5/0.2 & 10/0.3 & 15.0/0.22 \\
& GoMatch~\cite{zhou2022gomatch} & No & -- & ${\sim}$12\,MB & N/A & 25/0.6 & 283/8.1 & 48/4.8 & 335/9.9 & N/A \\
& HybridSC~\cite{camposeco2019hybrid} & No & -- & ${\sim}$1\,MB & N/A & 81/0.6 & 75/1.0 & 19/0.5 & 50/0.5 & N/A \\
\midrule
\multirow{2}{*}{\rotatebox{90}{APR}}
& PoseNet17~\cite{kendall2017geometric} & No & 4--24\,h & 50\,MB & 683/3.5 & 88/1.0 & 320/3.3 & 88/3.8 & 157/3.3 & 267.2/2.98 \\
& MS-Transformer~\cite{shavit2021mstransformer} & No & ${\sim}$7\,h & ${\sim}$18\,MB & N/A & 83/1.5 & 181/2.4 & 86/3.1 & 162/4.0 & N/A \\
\midrule
\multirow{3}{*}{\rotatebox{90}{SCR+D}}
& DSAC* (Full)~\cite{brachmann2022dsacstar} & Yes & 15\,h & 28\,MB & 49/0.3 & 15/0.3 & 21/0.4 & 5/0.3 & 13/0.4 & 20.6/0.34 \\
& SANet~\cite{yang2019sanet} & Yes & ${\sim}$1\,min & ${\sim}$260\,MB & 328/2.0 & 32/0.5 & 32/0.5 & 10/0.5 & 16/0.6 & 83.6/0.82 \\
& SRC~\cite{dong2022src} & Yes & 2\,min$^{\dagger}$ & 40\,MB & 81/0.5 & 39/0.7 & 38/0.5 & 19/1.0 & 31/1.0 & 41.6/0.74 \\
\midrule
\multirow{7}{*}{\rotatebox{90}{SCR (RGB only)}}
& DSAC* (Full)~\cite{brachmann2022dsacstar} & No & 15\,h & 28\,MB & 34/0.2 & 18/0.3 & 21/0.4 & 5/0.3 & 15/0.6 & 18.6/0.36 \\
& DSAC* (Tiny)~\cite{brachmann2022dsacstar} & No & 11\,h & 4\,MB & 98/0.5 & 27/0.4 & 33/0.6 & 11/0.5 & 56/1.8 & 45.0/0.76 \\
& ACE~\cite{brachmann2023ace} & No & 5\,min & 4\,MB & 43/0.2 & 28/0.4 & 31/0.6 & 5/0.3 & 18/0.6 & 25.0/0.42 \\
& Poker~\cite{brachmann2023ace} & No & 20\,min & 16\,MB & 28/0.1 & 18/0.3 & 25/0.5 & 5/0.3 & 9/0.3 & 17.0/0.30 \\
& GLACE~\cite{wang2024glace} & No & 80\,min & 13\,MB & \textbf{19}/0.1 & 19/0.3 & \textbf{17}/0.4 & \textbf{4}/0.2 & 9/0.3 & 13.6/0.26 \\
& MACE~\cite{liu2025mace} & No & 30\,min$^{\ddagger}$ & 5\,MB$^{\S}$ & 24/0.2 & \textbf{15}/0.3 & 19/0.4 & 5/0.2 & 9/0.3 & 14.4/0.28 \\
& DAC-MACE (ours) & No & 35\,min$^{\ddagger}$ & 10\,MB$^{\S}$ & \textbf{19}/0.1 & \textbf{15}/0.3 & 18/0.4 & 5/0.2 & \textbf{8}/0.3 & \textbf{13.0}/0.26 \\
\bottomrule
\end{tabular*}
\end{table*}

\begin{figure*}[t]
\centering
\includegraphics[width=\textwidth]{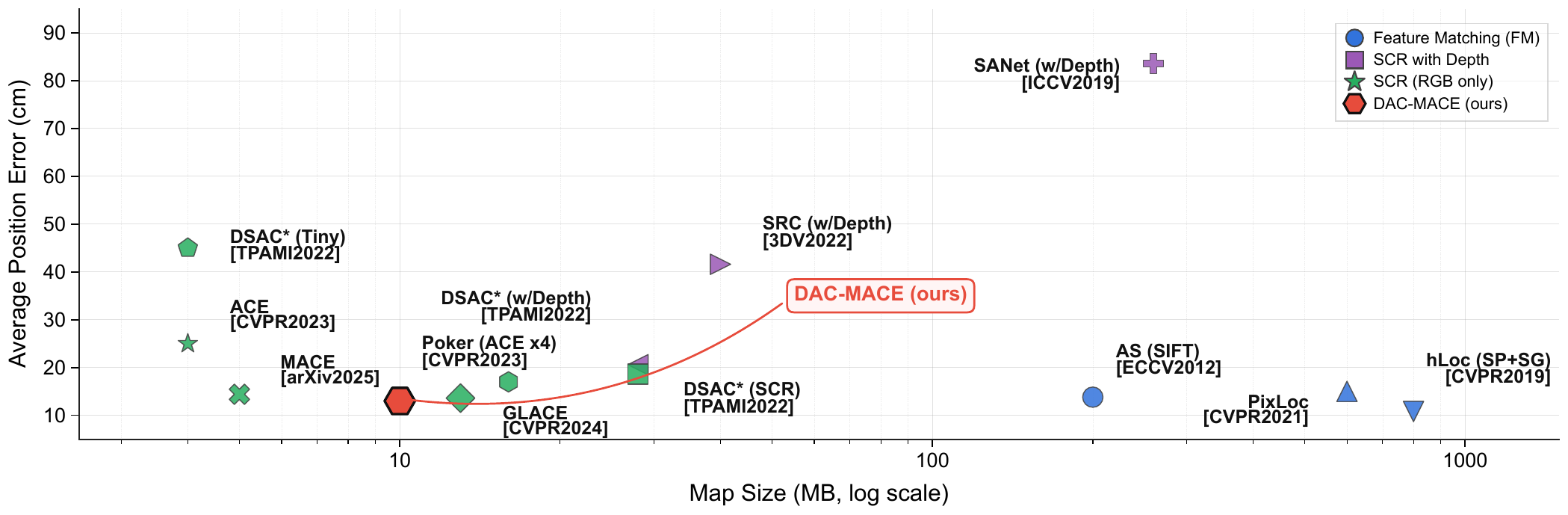}
\caption{Memory--accuracy trade-off on Cambridge Landmarks: map size versus the mean of per-scene median translation errors (values of Table~\ref{tab:cambridge}). DAC-MACE has the lowest error among the RGB-only SCR methods. Plotted method/value pairs: AS (200\,MB, 13.8\,cm), hLoc (800, 10.8), PixLoc (600, 15.0), DSAC* with depth (28, 20.6), SRC with depth (40, 41.6), SANet with depth (260, 83.6), RGB-only DSAC* (28, 18.6), DSAC* Tiny (4, 45.0), ACE (4, 25.0), Poker (16, 17.0), GLACE (13, 13.6), MACE (5, 14.4), and DAC-MACE (10, 13.0).}
\label{fig:mapsize}
\end{figure*}

\subsection{Dual-Threshold Pose Estimation}
\label{sec:pose}

Coordinate accuracy varies across scenes and images: a strict inlier threshold yields precise hypotheses when coordinates are accurate, whereas a loose threshold retains enough support when they are noisy. Given the dense prediction $\{(\p,\y_{\p})\}$, we therefore run two RANSAC searches with thresholds $\tau_{\mathrm{s}}<\tau_{\mathrm{l}}$, each with 256 hypotheses from minimal PnP samples and soft-inlier scoring as in DSAC*~\cite{brachmann2022dsacstar}, yielding $\Pose_{\mathrm{s}}$ and $\Pose_{\mathrm{l}}$. A looser threshold admits more inliers by construction, so the two hypotheses are compared by the threshold-normalized support
\begin{equation}
\rho(\Pose,\tau)=|\mathcal{I}(\Pose,\tau)|/\tau,
\label{eq:support}
\end{equation}
where $\mathcal{I}(\Pose,\tau)$ is the set of correspondences with reprojection error below $\tau$, and the hypothesis with the larger support is selected. The rule selects the loose hypothesis only if $|\mathcal{I}(\Pose_{\mathrm{l}},\tau_{\mathrm{l}})|>(\tau_{\mathrm{l}}/\tau_{\mathrm{s}})\,|\mathcal{I}(\Pose_{\mathrm{s}},\tau_{\mathrm{s}})|$, i.e., if it gathers at least $\tau_{\mathrm{l}}/\tau_{\mathrm{s}}=1.67$ times as many inliers. The rule does not assume that inlier counts grow linearly with the threshold; it fixes the exchange rate between precision and support so that the strict hypothesis is preferred unless the loose one is substantially better supported. The selected pose is refined by Levenberg--Marquardt (LM)~\cite{marquardt1963lm} on the reprojection error, first over inliers at 12\,px and then at 5\,px, with at least 50 correspondences required before each round. Like LO-RANSAC~\cite{chum2003lo}, the procedure combines hypothesis search with local optimization; unlike it, the two operating points are searched independently and the local optimization is applied only to the selected hypothesis. The constants of both branches are listed in Table~\ref{tab:hyper}.

% =====================================================================
\section{Experiments}
\label{sec:exp}

\subsection{Datasets, Reference Poses, and Metrics}

Cambridge Landmarks~\cite{kendall2015posenet} contains five outdoor scenes with SfM reference poses; the numbers of mapping and query images are listed in Table~\ref{tab:split}. Images are rescaled to a height of 480\,px. 7-Scenes~\cite{shotton2013scene} contains seven indoor scenes (Chess, Fire, Heads, Office, Pumpkin, Red Kitchen, and Stairs) recorded at $640{\times}480$. For 7-Scenes we use the original D-SLAM (KinectFusion) reference poses for all methods in Table~\ref{tab:7scenes}; methods evaluated against SfM pseudo ground truth are not directly comparable~\cite{brachmann2021limits}. We report per-scene median translation and rotation errors. For Cambridge, ``Avg.'' is the arithmetic mean of the five per-scene medians, not a pooled median, and is given with one decimal because the differences between the strongest methods are about 1\,cm; for 7-Scenes we also report the fraction of queries within (5\,cm,\,5$^\circ$).

\begin{table*}[t]
\centering
\caption{Cumulative Component Analysis on Cambridge Landmarks: Median Translation (cm) / Rotation ($^\circ$). Each Row Adds One Component to the Previous Row. The First Two Rows Are the Poker and MACE Configurations With the Values Reported in~\cite{brachmann2023ace,liu2025mace}; the GCE, RCA, and Dual-Threshold Rows Are the Reference Run (Seed 2089); the Last Row Is the Three-Seed Mean of the Full Model. $\Delta$Avg. Is the Change in Mean Translation Error With Respect to the Previous Row.}
\label{tab:ablation}
\footnotesize
\begin{tabular*}{\textwidth}{@{\extracolsep{\fill}}l ccccc cc@{}}
\toprule
Variant & Great Court & King's College & Old Hospital & Shop Facade & St Mary's & Avg. & $\Delta$Avg. \\
\midrule
ACE Poker~\cite{brachmann2023ace} & 28/0.1 & 18/0.3 & 25/0.5 & 5/0.3 & 9/0.3 & 17.0/0.30 & -- \\
+ MoE (MACE configuration)~\cite{liu2025mace} & 24/0.2 & 15/0.3 & 19/0.4 & 5/0.2 & 9/0.3 & 14.4/0.28 & $-2.6$ \\
+ GCE & 21/0.1 & 16/0.3 & 19/0.4 & 5/0.2 & 9/0.3 & 14.0/0.26 & $-0.4$ \\
+ RCA & 22/0.1 & 15/0.3 & 18/0.4 & 5/0.2 & 8/0.3 & 13.6/0.26 & $-0.4$ \\
+ Dual-threshold pose stage & 19/0.1 & 15/0.3 & 18/0.4 & 5/0.2 & 8/0.4 & 13.0/0.28 & $-0.6$ \\
\midrule
DAC-MACE, mean of three seeds & 19/0.1 & 15/0.3 & 18/0.4 & 5/0.2 & 8/0.3 & 13.0/0.26 & \\
\bottomrule
\end{tabular*}
\end{table*}

\subsection{Implementation Details}

\begin{table}[t]
\centering
\caption{Training and Pose-Estimation Settings, Shared by All Scenes}
\label{tab:hyper}
\footnotesize
\begin{tabularx}{\columnwidth}{@{}lX@{}}
\toprule
Item & Setting \\
\midrule
\multicolumn{2}{@{}l}{\emph{Mapping data}} \\
Augmentation & grayscale; brightness/contrast $\pm10\%$; rotation $\pm15^\circ$; scale $[0.67,1.5]$ \\
Buffer & 64\,M features, 1{,}024 per image, 16-bit \\
Batch & 163{,}840 features ($320\times16\times32$) \\
\multicolumn{2}{@{}l}{\emph{Optimization}} \\
Schedule & 16 epochs per expert, sequential over experts \\
Optimizer & AdamW, weight decay $10^{-4}$, no gradient clipping \\
Peak learning rate & $5{\times}10^{-3}$ (expert, RCA, GCE); $5{\times}10^{-4}$ (router); one-cycle \\
Reprojection loss & dynamic tanh, $\omega_{\max}=50$\,px, $\omega_{\min}=1$\,px \\
Load balancing & $\lambda=0.01$; bias EMA momentum 0.9, step 0.1 \\
Seeds & 2089, 2090, 2091 \\
\multicolumn{2}{@{}l}{\emph{Network}} \\
RCA & reduction 16, $7{\times}7$ kernel, $\gamma_{\mathrm{r}}^{(0)}=0.01$ \\
GCE & 8 heads, scales $\{1,2\}$, attention dropout 0.1, stochastic depth $\leq0.1$, LayerScale 0.01 \\
Router & MLP 512--512--256, LayerNorm, dropout 0.05, Gumbel $\tau=1$ \\
Experts & $K\in\{2,4\}$, $M=50$ centers each \\
\multicolumn{2}{@{}l}{\emph{Pose estimation}} \\
Strict branch & 256 hypotheses, $\tau_{\mathrm{s}}=6$\,px, sharpness 140, clamp 80\,px \\
Loose branch & 256 hypotheses, $\tau_{\mathrm{l}}=10$\,px, sharpness 100, clamp 100\,px \\
LM refinement & 12\,px then 5\,px, $\geq50$ correspondences, stride 2, $\leq12{,}000$ points \\
\bottomrule
\end{tabularx}
\end{table}

DAC-MACE is implemented on top of ACE~\cite{brachmann2023ace}; all settings are listed in Table~\ref{tab:hyper}. The Cambridge results of DAC-MACE are means over three training seeds; per-scene standard deviations are below 0.5\,cm and $0.05^\circ$. All models are trained on a single NVIDIA GPU with mixed precision. Following MACE~\cite{liu2025mace}, the mapping time is the wall-clock time of the routing stage, from the first buffer write to the saved model, and excludes the ACE pretraining of the cluster experts; the map size counts the expert heads in 16-bit precision, with 1\,MB $=2^{20}$\,bytes. The context modules (RCA, GCE, and the router) add 3.32\,M scene-specific parameters, which are reported separately in Table~\ref{tab:complexity}. Baseline accuracies are those reported in the original papers.

\begin{table*}[t]
\centering
\caption{7-Scenes With D-SLAM Reference Poses: Median Translation (cm) / Rotation ($^\circ$) and Recall@(5\,cm,\,5$^\circ$) in \%. ``SCR+D'' Denotes SCR Methods Trained With Depth or Geometry. ACE and ACE+GCE Were Trained by Us With Identical Settings; ACE+GCE Has a Single Head and No MoE. The Oracle Recall Is Reproduced From~\cite{sarlin2021pixloc}. Bold: Best Rotation Error and Recall per Column, and Best Translation Error Where It Is Unique at 1-cm Resolution.}
\label{tab:7scenes}
\footnotesize
\begin{tabular*}{\textwidth}{@{\extracolsep{\fill}}ll ccccccc c@{}}
\toprule
& Method & Chess & Fire & Heads & Office & Pumpkin & Red Kitchen & Stairs & Recall (\%) \\
\midrule
\multirow{8}{*}{\rotatebox{90}{FM}}
& DenseVLAD~\cite{torii2015place} & 21/12.5 & 33/13.8 & 15/14.9 & 28/11.2 & 31/11.3 & 30/12.3 & 25/15.8 & N/A \\
& Oracle~\cite{sarlin2021pixloc} & 16/12.3 & 26/13.6 & 12/14.7 & 20/11.5 & 19/14.0 & 18/15.0 & 17/18.1 & 0.17 \\
& AS~\cite{sattler2012improving} & 3/0.87 & 2/1.01 & 1/0.82 & 4/1.15 & 7/1.69 & 5/1.72 & 4/1.01 & 68.7 \\
& InLoc~\cite{taira2018inloc} & 3/1.05 & 3/1.07 & 2/1.16 & 3/1.05 & 5/1.55 & 4/1.31 & 9/2.47 & 66.3 \\
& hLoc~\cite{sarlin2019hloc,sarlin2020superglue} & 2/0.85 & 2/0.94 & 1/0.75 & 3/0.92 & 5/1.30 & 4/1.40 & 5/1.47 & 73.1 \\
& PixLoc~\cite{sarlin2021pixloc} & 2/0.80 & 2/0.73 & 1/0.82 & 3/0.82 & 4/1.21 & 3/1.20 & 5/1.30 & 75.7 \\
& PixLoc + oracle prior~\cite{sarlin2021pixloc} & 2/0.80 & 2/\textbf{0.70} & 1/0.78 & 3/\textbf{0.80} & 4/1.13 & 3/1.14 & 4/1.08 & 81.7 \\
& CAMNet~\cite{ding2019camnet} & 4/1.73 & 3/1.74 & 5/1.98 & 4/1.62 & 4/1.64 & 4/1.63 & 4/1.51 & N/A \\
\midrule
\multirow{3}{*}{\rotatebox{90}{SCR+D}}
& DSAC*~\cite{brachmann2022dsacstar} & 2/1.10 & 2/1.24 & 1/1.82 & 3/1.15 & 4/1.34 & 4/1.68 & 3/1.16 & 85.2 \\
& HSCNet~\cite{li2020hscnet} & 2/0.70 & 2/0.90 & 1/0.90 & 3/\textbf{0.80} & 4/\textbf{1.00} & 4/1.20 & 3/\textbf{0.80} & 84.8 \\
& SANet~\cite{yang2019sanet} & 3/0.88 & 3/1.08 & 2/1.48 & 3/1.00 & 5/1.32 & 4/1.40 & 16/4.59 & 68.2 \\
\midrule
\multirow{2}{*}{\rotatebox{90}{SCR}}
& ACE~\cite{brachmann2023ace} (our run) & 2/0.71 & 2/0.81 & 1/0.68 & 3/0.86 & 4/1.11 & 4/1.31 & 4/1.07 & 86.56 \\
& ACE + GCE (ours) & 2/\textbf{0.55} & 2/0.76 & 1/\textbf{0.65} & \textbf{2}/0.83 & \textbf{3}/1.14 & 3/\textbf{1.05} & 13/2.50 & \textbf{87.60} \\
\bottomrule
\end{tabular*}
\end{table*}

\subsection{Comparison on Cambridge Landmarks}
\label{sec:cambridge}

Table~\ref{tab:cambridge} compares DAC-MACE with feature-matching (FM), absolute pose regression (APR), and SCR methods. Among the RGB-only SCR methods, DAC-MACE has the lowest mean translation error, 13.0\,cm, against 13.6\,cm for GLACE and 14.4\,cm for MACE, and it is the best or tied best on Great Court, King's College, and St Mary's Church. The largest improvement is on Great Court, the largest scene, where the error falls from 24\,cm (MACE) to 19\,cm, matching GLACE; GLACE remains better on Old Hospital and Shop Facade by 1\,cm. Relative to the RGB-only DSAC* (18.6\,cm), whose mapping takes 15\,h, DAC-MACE lowers the mean error by 5.6\,cm with 35\,min of mapping. The ranking does not depend on the pose stage: with the single-threshold stage of ACE and MACE, DAC-MACE averages 13.4\,cm (Table~\ref{tab:pose}), still below GLACE and MACE.

Fig.~\ref{fig:mapsize} places the methods in the map-size--accuracy plane. The expert heads of DAC-MACE occupy 10\,MB for $K{=}4$, between MACE (5\,MB) and GLACE (13\,MB) and 20--80 times less than the structure-based methods, which reach 10.8--15.0\,cm with 200--800\,MB. Among the compact representations below 50\,MB, DAC-MACE gives the lowest error.

\subsection{Qualitative Trajectory Analysis}
\label{sec:trajectory}

\begin{figure}[t]
\centering
\includegraphics[width=\columnwidth]{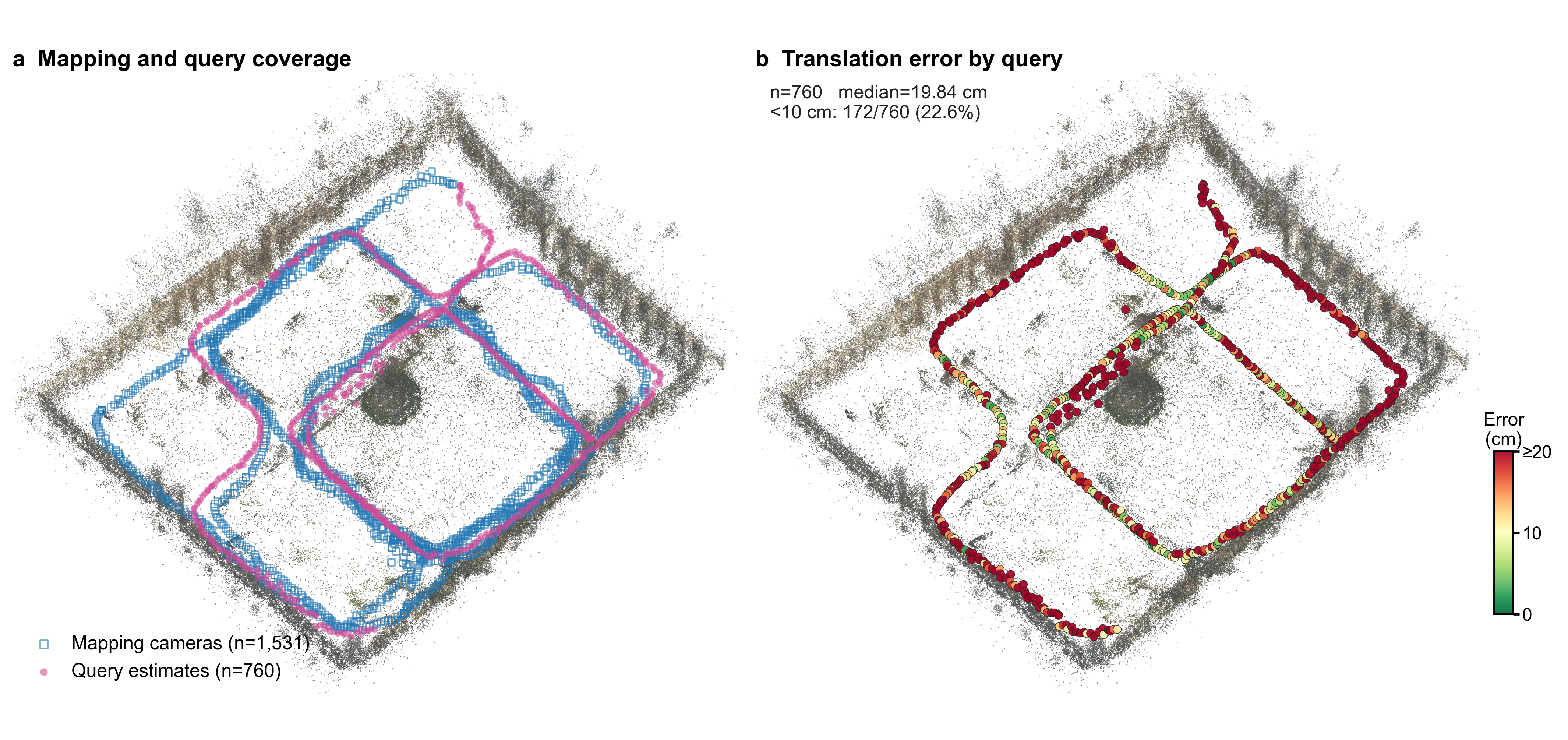}
\caption{Trajectory visualization on Great Court. (a) Mapping-camera centers (blue squares) and all 760 estimated query camera centers (pink) over the sparse SfM point cloud. (b) The same query estimates colored by translation error; values of 20\,cm and above saturate the color scale. The point cloud is thinned for display, while every query is shown. Camera centers are drawn without frustums, so the visualization does not depend on a camera-forward convention.}
\label{fig:greatcourt}
\end{figure}

\begin{figure}[t]
\centering
\includegraphics[width=\columnwidth]{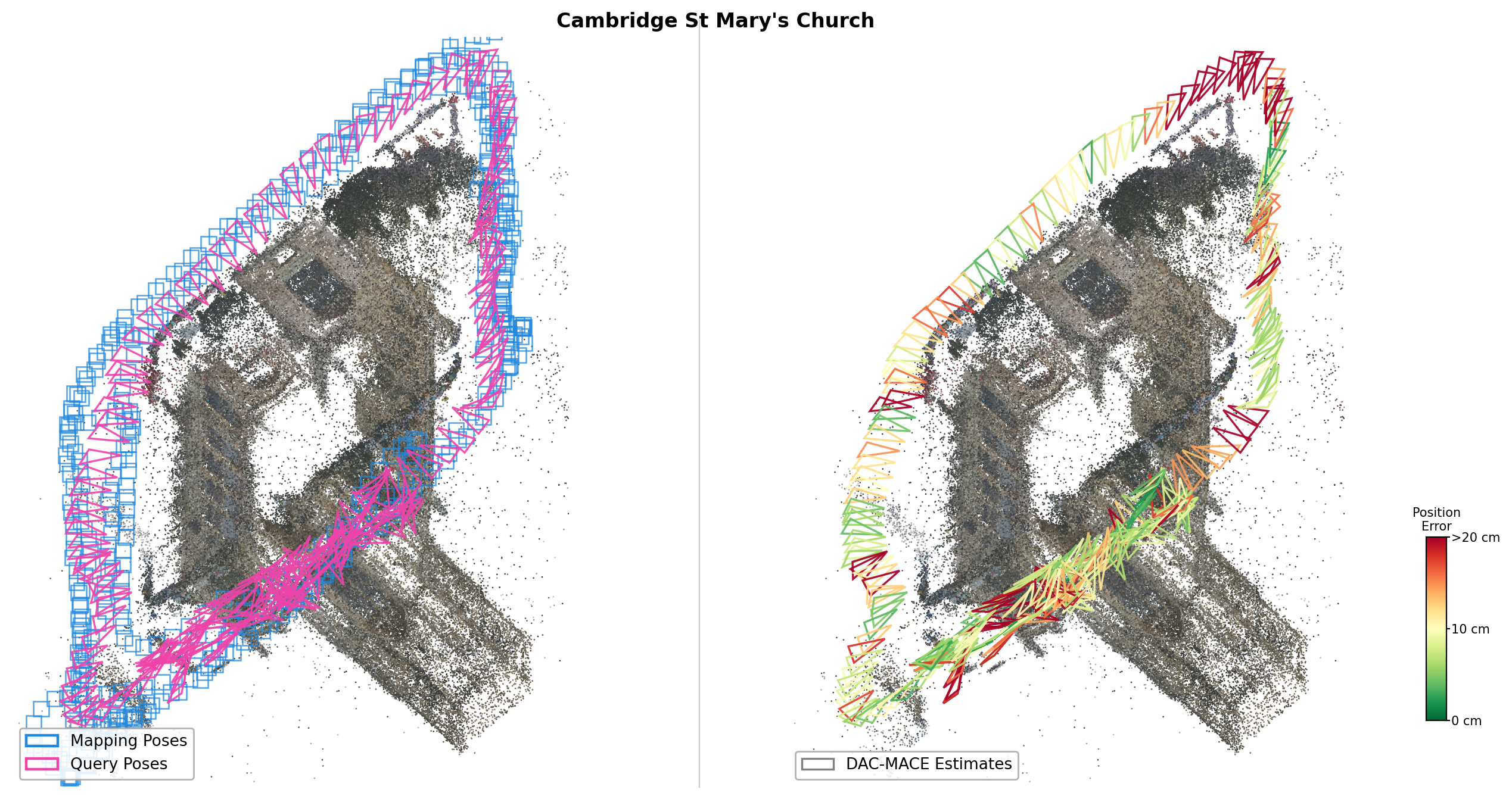}
\caption{Trajectory visualization on St Mary's Church. Left: mapping and estimated query camera positions over the SfM point cloud. Right: estimated query positions colored by translation error.}
\label{fig:stmarys}
\end{figure}

Figs.~\ref{fig:greatcourt} and~\ref{fig:stmarys} show where in the scene DAC-MACE localizes the query images. On Great Court, Fig.~\ref{fig:greatcourt}b shows how the translation errors of the reference run are distributed over the court; 172 of the 760 queries (22.6\%) are localized within 10\,cm. Fig.~\ref{fig:stmarys} gives the corresponding visualization for St Mary's Church, whose long facade contains repeated windows. These visualizations localize failures spatially; the benchmark medians in Table~\ref{tab:cambridge} remain the quantitative result.

\subsection{Component Analysis}
\label{sec:ablation}

\begin{table}[t]
\centering
\caption{Per-Scene Change in Median Translation Error (cm) Contributed by Each Component, Derived From Consecutive Rows of Table~\ref{tab:ablation}. Negative Values Denote Improvement.}
\label{tab:incremental}
\footnotesize
\begin{tabular*}{\columnwidth}{@{\extracolsep{\fill}}l cccc@{}}
\toprule
Scene & MoE & GCE & RCA & Dual-threshold \\
\midrule
Great Court & $-4$ & $-3$ & $+1$ & $-3$ \\
King's College & $-3$ & $+1$ & $-1$ & 0 \\
Old Hospital & $-6$ & 0 & $-1$ & 0 \\
Shop Facade & 0 & 0 & 0 & 0 \\
St Mary's Church & 0 & 0 & $-1$ & 0 \\
\midrule
Avg. & $-2.6$ & $-0.4$ & $-0.4$ & $-0.6$ \\
\bottomrule
\end{tabular*}
\end{table}

Table~\ref{tab:ablation} adds the components one at a time, and Table~\ref{tab:incremental} lists the per-scene change contributed by each. Replacing the fixed Poker ensemble with the learned MoE of MACE gives the largest single reduction, from 17.0 to 14.4\,cm, with gains on Great Court, King's College, and Old Hospital. Conditioning the router on GCE features lowers the mean to 14.0\,cm; the gain comes from Great Court, the largest and most repetitive scene, where the error drops from 24 to 21\,cm, while King's College changes by $+1$\,cm. Adding RCA lowers the mean to 13.6\,cm through King's College, Old Hospital, and St Mary's Church, two of which are dominated by repeated window patterns, at the cost of 1\,cm on Great Court. The dual-threshold pose stage then brings Great Court to 19\,cm and the mean to 13.0\,cm. Every component therefore lowers the mean error, by 0.4--0.6\,cm for each of the three proposed components and 0.8\,cm for RCA and GCE together, and the per-scene effects are complementary: GCE acts mainly on the largest scene and RCA on facade scenes. As the intermediate rows come from a single seed at 1-cm resolution, per-scene changes of 1\,cm are indicative only; the five-scene mean, which the three-seed run of the full model reproduces, is the more reliable quantity.

\subsection{Pose-Stage Analysis}
\label{sec:pose_exp}

\begin{table}[t]
\centering
\caption{Pose-Stage Ablation on the Fixed Trained DAC-MACE Models (Mean of Three Seeds): Median Translation (cm) / Rotation ($^\circ$). Only the Inference-Time Pose Stage Is Changed.}
\label{tab:pose}
\footnotesize
\begin{tabular*}{\columnwidth}{@{\extracolsep{\fill}}l cc@{}}
\toprule
Scene & \begin{tabular}[c]{@{}c@{}}Single-threshold\\PnP+LM\end{tabular} & \begin{tabular}[c]{@{}c@{}}Dual-threshold\\RANSAC+LM\end{tabular} \\
\midrule
Great Court & 20/0.1 & 19/0.1 \\
King's College & 16/0.3 & 15/0.3 \\
Old Hospital & 18/0.4 & 18/0.4 \\
Shop Facade & 5/0.2 & 5/0.2 \\
St Mary's Church & 8/0.3 & 8/0.3 \\
\midrule
Avg. & 13.4/0.26 & 13.0/0.26 \\
\bottomrule
\end{tabular*}
\end{table}

\begin{table}[t]
\centering
\caption{Cambridge Data Split and Dual-Threshold Branch Selection. Strict and Loose Counts Sum to Each Scene's Number of Query Images.}
\label{tab:split}
\footnotesize
\begin{tabular*}{\columnwidth}{@{\extracolsep{\fill}}l c cc c@{}}
\toprule
Scene & Mapping & Strict, 6\,px & Loose, 10\,px & Query \\
\midrule
Great Court & 1{,}531 & 617 (81.2\%) & 143 (18.8\%) & 760 \\
King's College & 1{,}220 & 343 (100.0\%) & 0 (0.0\%) & 343 \\
Old Hospital & 895 & 179 (98.4\%) & 3 (1.6\%) & 182 \\
Shop Facade & 231 & 3 (2.9\%) & 100 (97.1\%) & 103 \\
St Mary's Church & 1{,}487 & 448 (84.5\%) & 82 (15.5\%) & 530 \\
\midrule
All scenes & 5{,}364 & 1{,}590 (82.9\%) & 328 (17.1\%) & 1{,}918 \\
\bottomrule
\end{tabular*}
\end{table}

Table~\ref{tab:pose} keeps the trained networks fixed and changes only the pose stage, from the single-threshold stage of ACE and MACE (one RANSAC search at 10\,px followed by LM refinement) to the dual-threshold stage. The mean translation error falls from 13.4 to 13.0\,cm, with improvements on Great Court (20 to 19\,cm) and King's College (16 to 15\,cm) and no change on the other scenes at the displayed precision. Table~\ref{tab:split} shows how often each branch is selected. Over all 1{,}918 Cambridge queries, the strict branch is selected for 82.9\% and the loose branch for 17.1\%, and the split depends strongly on the scene. On King's College and Old Hospital the strict branch is selected almost always, so the gain on King's College comes from the strict 6-px operating point and the two-round refinement. On Great Court, about one query in five is better supported at the loose threshold. On Shop Facade, the smallest scene, 97.1\% of the queries select the loose branch without loss of accuracy, which indicates that a single threshold is not optimal for all scenes.

\subsection{Routing Analysis}
\label{sec:routing}

\begin{figure}[t]
\centering
\includegraphics[width=\columnwidth]{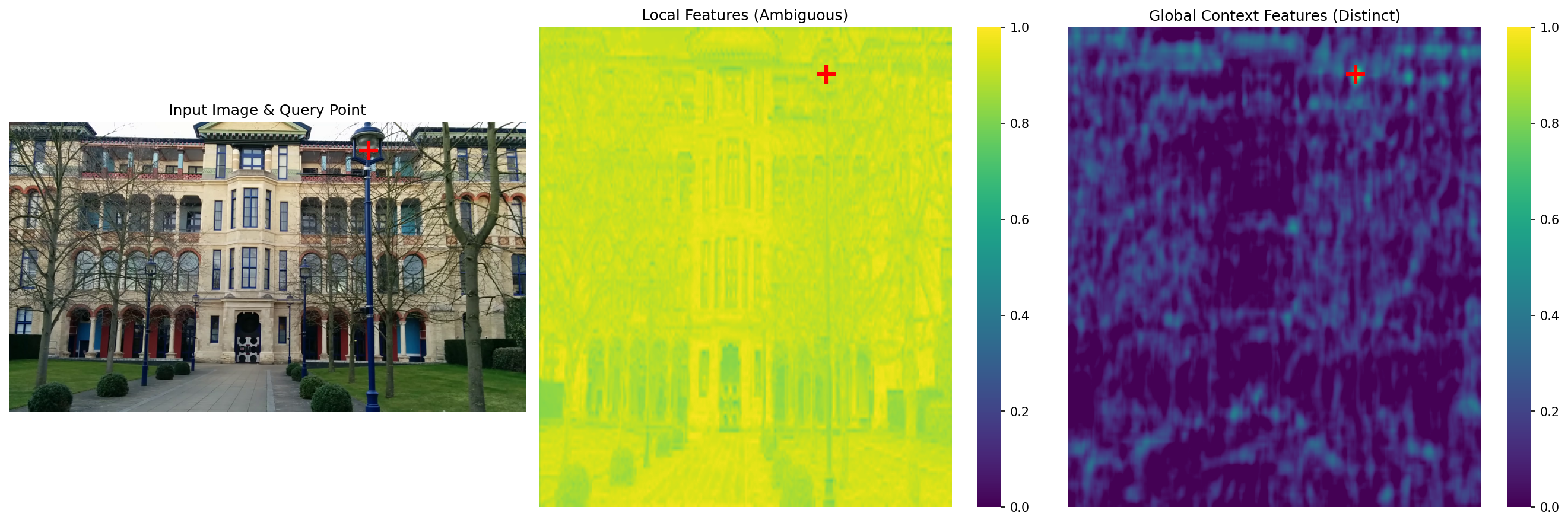}
\caption{Qualitative feature-response diagnostic on a repetitive Old Hospital facade. Left: query image; the red cross marks the query location. Middle: cosine similarity between the query feature and every feature of the image before GCE. Right: the same after GCE. Each map is min--max normalized to $[0,1]$ independently and bilinearly upsampled, so colors indicate spatial concentration rather than calibrated confidence.}
\label{fig:ambiguity}
\end{figure}

\begin{figure}[t]
\centering
\includegraphics[width=\columnwidth]{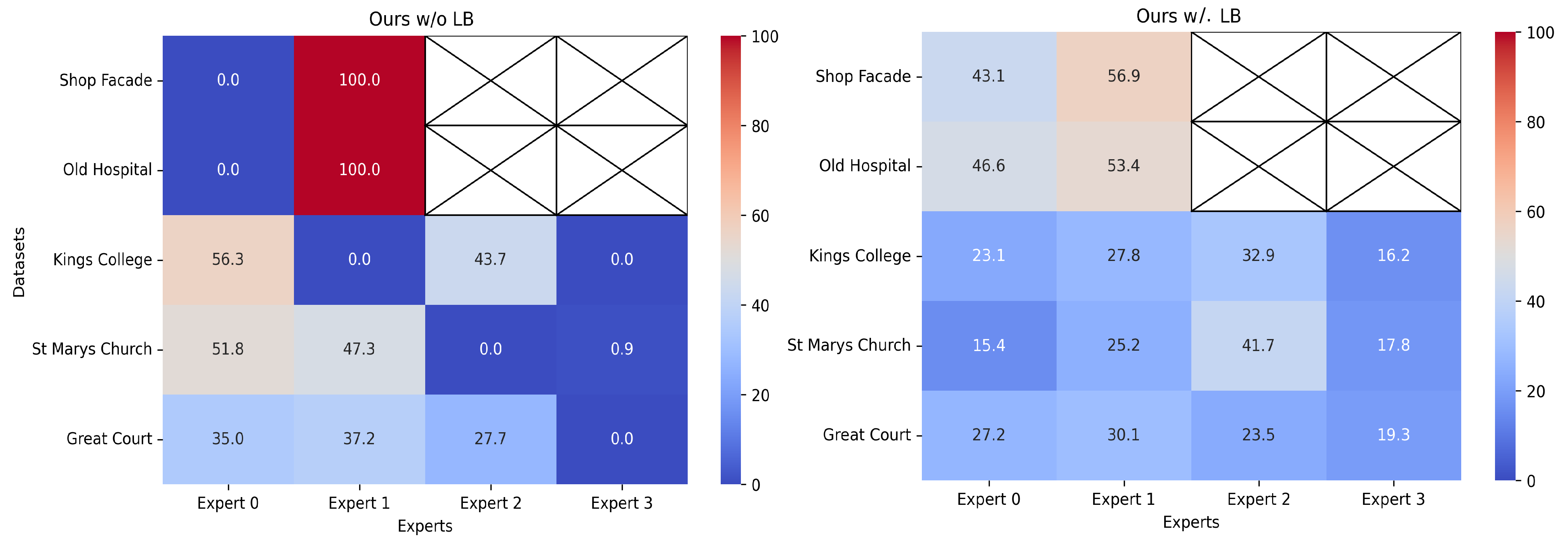}
\caption{Hard expert-selection ratios (\%) of DAC-MACE on Cambridge Landmarks, without (left) and with (right) load balancing, i.e., the standard-deviation penalty in~\eqref{eq:loss} and the router-bias update. Crossed cells mark experts that do not exist in the $K{=}2$ models.}
\label{fig:experts}
\end{figure}

\begin{table}[t]
\centering
\caption{Expert-Usage Statistics Derived From Fig.~\ref{fig:experts}, Without/With Load Balancing (LB). With Hard Selection Ratios $p_k$, $H_{\mathrm{norm}}=-\sum_k p_k\log p_k/\log K$ and $N_{\mathrm{eff}}=\exp(-\sum_k p_k\log p_k)$; Max: Largest Expert Share (\%).}
\label{tab:routing}
\footnotesize
\begin{tabular*}{\columnwidth}{@{\extracolsep{\fill}}l c cc cc cc@{}}
\toprule
& & \multicolumn{2}{c}{$H_{\mathrm{norm}}$} & \multicolumn{2}{c}{$N_{\mathrm{eff}}$} & \multicolumn{2}{c}{Max (\%)} \\
\cmidrule(lr){3-4}\cmidrule(lr){5-6}\cmidrule(lr){7-8}
Scene & $K$ & w/o & w/ & w/o & w/ & w/o & w/ \\
\midrule
Shop Facade & 2 & 0.000 & 0.986 & 1.00 & 1.98 & 100.0 & 56.9 \\
Old Hospital & 2 & 0.000 & 0.997 & 1.00 & 2.00 & 100.0 & 53.4 \\
King's College & 4 & 0.494 & 0.977 & 1.98 & 3.88 & 56.3 & 32.9 \\
St Mary's Church & 4 & 0.532 & 0.943 & 2.09 & 3.70 & 51.8 & 41.7 \\
Great Court & 4 & 0.787 & 0.990 & 2.98 & 3.95 & 37.2 & 30.1 \\
\bottomrule
\end{tabular*}
\end{table}

\color{black}

The router is analyzed with respect to the two failure modes of Section~\ref{sec:intro}.

\emph{Perceptual routing ambiguity.} For a fixed query location on a repetitive Old Hospital facade, we compute the cosine similarity between its feature and every feature of the image, before and after GCE (Fig.~\ref{fig:ambiguity}). Before GCE, the response is diffuse: features across the repeated facade are similar to the query feature, so a router operating on them cannot tell the repetitions apart. After GCE, the response is spatially more concentrated. This qualitative diagnostic illustrates how context mixing changes the router input; its effect on accuracy is given by the component analysis in Table~\ref{tab:ablation}.

\emph{Expert-load collapse.} Fig.~\ref{fig:experts} and Table~\ref{tab:routing} report hard expert-selection ratios and three statistics derived from them. Without load balancing, the $K{=}2$ models route every pixel to a single expert, and the $K{=}4$ models effectively use two or three experts ($N_{\mathrm{eff}}$ of 1.98--2.98). With balancing, the normalized entropy exceeds 0.94 in every scene, the effective number of experts reaches 3.70--3.95 of 4 for the large scenes, and the largest expert share falls in every scene. These statistics quantify how evenly the capacity is used, i.e., expert-load collapse rather than routing correctness.

\subsection{Model Complexity}
\label{sec:complexity}

\begin{table}[t]
\centering
\caption{Parameters and Floating-Point Operations of DAC-MACE ($K{=}4$) for a $480{\times}853$ Cambridge Query ($60{\times}107$ Features), Computed Analytically From the Network Definition. Last Column: Scene-Specific Parameters (M).}
\label{tab:complexity}
\footnotesize
\begin{tabular*}{\columnwidth}{@{\extracolsep{\fill}}l c c c@{}}
\toprule
Component & Params (M) & GFLOPs & Scene (M) \\
\midrule
ACE encoder (frozen) & 5.57 & 78.1 & -- \\
RCA & 0.03 & ${<}0.1$ & 0.03 \\
GCE, scale-1 branch & 1.05 & 97.9 & 1.05 \\
GCE, scale-2 branch & 1.05 & 8.5 & 1.05 \\
GCE, fusion & 0.53 & 6.7 & 0.53 \\
Router & 0.66 & 8.4 & 0.66 \\
Experts ($4\times$) & 5.36 & 68.7 & 5.36 \\
\midrule
DAC-MACE total & 14.25 & 268.3 & 8.68 \\
ACE (encoder + one head) & 7.67 & 105.0 & 2.10 \\
\bottomrule
\end{tabular*}
\end{table}

\color{black}

Table~\ref{tab:complexity} lists the parameters and operations of each component, counted from the network definition. Besides the four expert heads (5.36\,M parameters, 10\,MB in 16-bit precision), the context modules add 3.32\,M scene-specific parameters (6.3\,MB). Per query, DAC-MACE needs about 2.6 times the operations of ACE. Scale-1 attention over the 6{,}420 feature tokens accounts for 98\,GFLOPs, more than all four experts together; its attention matrices occupy about 0.6\,GB in 16-bit precision unless a memory-efficient attention kernel is used. The scale-2 branch costs less than a tenth of the scale-1 branch, so restricting GCE to the pooled scale, or using windowed or linear attention, is the most direct way to reduce the query cost for embedded deployment.

\subsection{Results on 7-Scenes}
\label{sec:7scenes}

7-Scenes is small and indoor, so we use it to test whether the context module also helps a single ACE head without MoE (Table~\ref{tab:7scenes}). Adding GCE to ACE raises the aggregate recall from 86.56\% to 87.60\%, the highest in the table and above the depth-supervised DSAC* (85.2\%) and HSCNet (84.8\%). It lowers the median translation error on Office, Pumpkin, and Red Kitchen (by 1\,cm each), lowers the median rotation error on Chess (0.71 to $0.55^\circ$) and Red Kitchen (1.31 to $1.05^\circ$), and keeps or improves the remaining errors except the Pumpkin rotation ($1.11$ to $1.14^\circ$) and Stairs. On Stairs, the median translation error rises from 4 to 13\,cm. Stairs consists largely of identical steps, and content-based attention without positional encoding can aggregate features of different steps that look alike. The context module therefore transfers to small indoor scenes, with repetitive geometry of near-identical appearance as its failure mode.

\subsection{Limitations}
\label{sec:limitations}

The full system is evaluated on Cambridge Landmarks, whose scenes span up to about 8{,}000\,m$^2$; city-scale benchmarks such as Aachen Day--Night~\cite{sattler2018benchmarking} and LaMAR~\cite{sarlin2022lamar}, together with a comparison to R-SCoRe~\cite{jiang2025rscore} on them, are left to future work, as are changing illumination and seasons. The component analysis is cumulative, and its intermediate rows come from a single seed. The dual-threshold stage was evaluated on DAC-MACE against the single-threshold stage of ACE, but it was not applied to the baselines or compared with LO-RANSAC~\cite{chum2003lo} or MAGSAC++~\cite{barath2020magsacpp}. Routing ambiguity is illustrated qualitatively (Fig.~\ref{fig:ambiguity}) rather than measured; its reduction is inferred from the accuracy changes in Table~\ref{tab:ablation}. RCA and GCE are trained on shuffled features (Section~\ref{sec:protocol}); training on image-coherent feature windows may let the spatial operators learn genuine image-space context. Finally, soft routing evaluates every expert, and the complexity in Table~\ref{tab:complexity} is computed analytically rather than measured on embedded hardware; sparse top-1 inference and lighter attention would reduce the query cost.

% =====================================================================
\section{Conclusion}
\label{sec:conclusion}

DAC-MACE conditions the expert routing of scene coordinate regression on recalibrated, context-mixed, and image-pooled features instead of on the local descriptor alone, and recovers the camera pose with a dual-threshold RANSAC--LM stage. On Cambridge Landmarks it obtains the lowest mean of per-scene median translation errors among the RGB-only SCR methods compared, 13.0\,cm, with 10\,MB of expert parameters and 35\,min of routing-stage mapping, and reduces the error on the largest scene from 24 to 19\,cm relative to MACE. Each proposed component lowers the mean error, GCE mainly on the largest scene and RCA on facade scenes, and the pose stage adapts its operating point to the scene. On 7-Scenes, the context module raises the recall of a single ACE head and lowers or keeps its median translation error on six of seven scenes. These properties make context-conditioned SCR a compact option for camera-based relocalization on mobile platforms; evaluation on city-scale and long-term benchmarks and a lighter attention design for embedded processors are the next steps.

\section*{Acknowledgment}
The authors used an AI writing assistant (Claude, Anthropic) for language polishing of this manuscript. All technical content, experimental data, and conclusions were produced and verified by the authors, who take full responsibility for the content of this article.

\bibliographystyle{IEEEtran}
\bibliography{refs}

\begin{thebibliography}{10}
\providecommand{\url}[1]{#1}
\csname url@samestyle\endcsname
\providecommand{\newblock}{\relax}
\providecommand{\bibinfo}[2]{#2}
\providecommand{\BIBentrySTDinterwordspacing}{\spaceskip=0pt\relax}
\providecommand{\BIBentryALTinterwordstretchfactor}{4}
\providecommand{\BIBentryALTinterwordspacing}{\spaceskip=\fontdimen2\font plus
\BIBentryALTinterwordstretchfactor\fontdimen3\font minus
  \fontdimen4\font\relax}
\providecommand{\BIBforeignlanguage}[2]{{%
\expandafter\ifx\csname l@#1\endcsname\relax
\typeout{** WARNING: IEEEtran.bst: No hyphenation pattern has been}%
\typeout{** loaded for the language `#1'. Using the pattern for}%
\typeout{** the default language instead.}%
\else
\language=\csname l@#1\endcsname
\fi
#2}}
\providecommand{\BIBdecl}{\relax}
\BIBdecl

\bibitem{kendall2015posenet}
A.~Kendall, M.~Grimes, and R.~Cipolla, ``{PoseNet}: A convolutional network for
  real-time 6-{DOF} camera relocalization,'' in \emph{Proc. IEEE Int. Conf.
  Comput. Vis. (ICCV)}, 2015, pp. 2938--2946.

\bibitem{sattler2018benchmarking}
T.~Sattler, W.~Maddern, C.~Toft, A.~Torii, L.~Hammarstrand, E.~Stenborg,
  D.~Safari, M.~Okutomi, M.~Pollefeys, J.~Sivic, F.~Kahl, and T.~Pajdla,
  ``Benchmarking {6DOF} outdoor visual localization in changing conditions,''
  in \emph{Proc. IEEE/CVF Conf. Comput. Vis. Pattern Recognit. (CVPR)}, 2018,
  pp. 8601--8610.

\bibitem{sattler2012improving}
T.~Sattler, B.~Leibe, and L.~Kobbelt, ``Improving image-based localization by
  active correspondence search,'' in \emph{Proc. Eur. Conf. Comput. Vis.
  (ECCV)}, 2012, pp. 752--765.

\bibitem{sarlin2019hloc}
P.-E. Sarlin, C.~Cadena, R.~Siegwart, and M.~Dymczyk, ``From coarse to fine:
  Robust hierarchical localization at large scale,'' in \emph{Proc. IEEE/CVF
  Conf. Comput. Vis. Pattern Recognit. (CVPR)}, 2019, pp. 12\,716--12\,725.

\bibitem{shotton2013scene}
J.~Shotton, B.~Glocker, C.~Zach, S.~Izadi, A.~Criminisi, and A.~Fitzgibbon,
  ``Scene coordinate regression forests for camera relocalization in {RGB-D}
  images,'' in \emph{Proc. IEEE Conf. Comput. Vis. Pattern Recognit. (CVPR)},
  2013, pp. 2930--2937.

\bibitem{brachmann2017dsac}
E.~Brachmann, A.~Krull, S.~Nowozin, J.~Shotton, F.~Michel, S.~Gumhold, and
  C.~Rother, ``{DSAC} -- differentiable {RANSAC} for camera localization,'' in
  \emph{Proc. IEEE Conf. Comput. Vis. Pattern Recognit. (CVPR)}, 2017, pp.
  6684--6692.

\bibitem{brachmann2022dsacstar}
E.~Brachmann and C.~Rother, ``Visual camera re-localization from {RGB} and
  {RGB-D} images using {DSAC},'' \emph{IEEE Trans. Pattern Anal. Mach.
  Intell.}, vol.~44, no.~9, pp. 5847--5865, 2022.

\bibitem{brachmann2023ace}
E.~Brachmann, T.~Cavallari, and V.~A. Prisacariu, ``Accelerated coordinate
  encoding: Learning to relocalize in minutes using {RGB} and poses,'' in
  \emph{Proc. IEEE/CVF Conf. Comput. Vis. Pattern Recognit. (CVPR)}, 2023, pp.
  5044--5053.

\bibitem{brachmann2019esac}
E.~Brachmann and C.~Rother, ``Expert sample consensus applied to camera
  re-localization,'' in \emph{Proc. IEEE/CVF Int. Conf. Comput. Vis. (ICCV)},
  2019, pp. 7525--7534.

\bibitem{liu2025mace}
M.~Liu, D.~Fan, H.~Que, H.~Gao, X.~Liu, S.~Peng, M.~Lin, S.~Gu, R.~Ye, W.~Qiu,
  H.~Yao, R.~Zhang, and X.~Huang, ``{MACE}: Mixture-of-experts accelerated
  coordinate encoding for large-scale scene localization and rendering,''
  arXiv:2510.14251, 2025.

\bibitem{torii2015place}
A.~Torii, R.~Arandjelovi{\'c}, J.~Sivic, M.~Okutomi, and T.~Pajdla, ``24/7
  place recognition by view synthesis,'' in \emph{Proc. IEEE Conf. Comput. Vis.
  Pattern Recognit. (CVPR)}, 2015, pp. 1808--1817.

\bibitem{shazeer2017moe}
N.~Shazeer, A.~Mirhoseini, K.~Maziarz, A.~Davis, Q.~Le, G.~Hinton, and J.~Dean,
  ``Outrageously large neural networks: The sparsely-gated mixture-of-experts
  layer,'' in \emph{Proc. Int. Conf. Learn. Represent. (ICLR)}, 2017.

\bibitem{fedus2022switch}
W.~Fedus, B.~Zoph, and N.~Shazeer, ``Switch transformers: Scaling to trillion
  parameter models with simple and efficient sparsity,'' \emph{J. Mach. Learn.
  Res.}, vol.~23, no. 120, pp. 1--39, 2022.

\bibitem{sarlin2020superglue}
P.-E. Sarlin, D.~DeTone, T.~Malisiewicz, and A.~Rabinovich, ``{SuperGlue}:
  Learning feature matching with graph neural networks,'' in \emph{Proc.
  IEEE/CVF Conf. Comput. Vis. Pattern Recognit. (CVPR)}, 2020, pp. 4938--4947.

\bibitem{arandjelovic2016netvlad}
R.~Arandjelovi{\'c}, P.~Gronat, A.~Torii, T.~Pajdla, and J.~Sivic, ``{NetVLAD}:
  {CNN} architecture for weakly supervised place recognition,'' in \emph{Proc.
  IEEE Conf. Comput. Vis. Pattern Recognit. (CVPR)}, 2016, pp. 5297--5307.

\bibitem{taira2018inloc}
H.~Taira, M.~Okutomi, T.~Sattler, M.~Cimpoi, M.~Pollefeys, J.~Sivic, T.~Pajdla,
  and A.~Torii, ``{InLoc}: Indoor visual localization with dense matching and
  view synthesis,'' in \emph{Proc. IEEE/CVF Conf. Comput. Vis. Pattern
  Recognit. (CVPR)}, 2018, pp. 7199--7209.

\bibitem{ding2019camnet}
M.~Ding, Z.~Wang, J.~Sun, J.~Shi, and P.~Luo, ``{CamNet}: Coarse-to-fine
  retrieval for camera re-localization,'' in \emph{Proc. IEEE/CVF Int. Conf.
  Comput. Vis. (ICCV)}, 2019, pp. 2871--2880.

\bibitem{sarlin2021pixloc}
P.-E. Sarlin, A.~Unagar, M.~Larsson, H.~Germain, C.~Toft, V.~Larsson,
  M.~Pollefeys, V.~Lepetit, L.~Hammarstrand, F.~Kahl, and T.~Sattler, ``Back to
  the feature: Learning robust camera localization from pixels to pose,'' in
  \emph{Proc. IEEE/CVF Conf. Comput. Vis. Pattern Recognit. (CVPR)}, 2021, pp.
  3247--3257.

\bibitem{camposeco2019hybrid}
F.~Camposeco, A.~Cohen, M.~Pollefeys, and T.~Sattler, ``Hybrid scene
  compression for visual localization,'' in \emph{Proc. IEEE/CVF Conf. Comput.
  Vis. Pattern Recognit. (CVPR)}, 2019, pp. 7653--7662.

\bibitem{zhou2022gomatch}
Q.~Zhou, S.~Agostinho, A.~O{\v{s}}ep, and L.~Leal-Taix{\'e}, ``Is geometry
  enough for matching in visual localization?'' in \emph{Proc. Eur. Conf.
  Comput. Vis. (ECCV)}, 2022, pp. 407--425.

\bibitem{kendall2017geometric}
A.~Kendall and R.~Cipolla, ``Geometric loss functions for camera pose
  regression with deep learning,'' in \emph{Proc. IEEE Conf. Comput. Vis.
  Pattern Recognit. (CVPR)}, 2017, pp. 5974--5983.

\bibitem{shavit2021mstransformer}
Y.~Shavit, R.~Ferens, and Y.~Keller, ``Learning multi-scene absolute pose
  regression with transformers,'' in \emph{Proc. IEEE/CVF Int. Conf. Comput.
  Vis. (ICCV)}, 2021, pp. 2733--2742.

\bibitem{brachmann2018learning}
E.~Brachmann and C.~Rother, ``Learning less is more -- {6D} camera localization
  via {3D} surface regression,'' in \emph{Proc. IEEE/CVF Conf. Comput. Vis.
  Pattern Recognit. (CVPR)}, 2018, pp. 4654--4662.

\bibitem{li2020hscnet}
X.~Li, S.~Wang, Y.~Zhao, J.~Verbeek, and J.~Kannala, ``Hierarchical scene
  coordinate classification and regression for visual localization,'' in
  \emph{Proc. IEEE/CVF Conf. Comput. Vis. Pattern Recognit. (CVPR)}, 2020, pp.
  11\,983--11\,992.

\bibitem{wang2024hscnetpp}
S.~Wang, Z.~Laskar, I.~Melekhov, X.~Li, Y.~Zhao, G.~Tolias, and J.~Kannala,
  ``{HSCNet++}: Hierarchical scene coordinate classification and regression for
  visual localization with transformer,'' \emph{Int. J. Comput. Vis.}, vol.
  132, no.~7, pp. 2530--2550, 2024.

\bibitem{dong2022src}
S.~Dong, S.~Wang, Y.~Zhuang, J.~Kannala, M.~Pollefeys, and B.~Chen, ``Visual
  localization via few-shot scene region classification,'' in \emph{Proc. Int.
  Conf. 3D Vis. (3DV)}, 2022, pp. 393--402.

\bibitem{yang2019sanet}
L.~Yang, Z.~Bai, C.~Tang, H.~Li, Y.~Furukawa, and P.~Tan, ``{SANet}: Scene
  agnostic network for camera localization,'' in \emph{Proc. IEEE/CVF Int.
  Conf. Comput. Vis. (ICCV)}, 2019, pp. 42--51.

\bibitem{wang2024glace}
F.~Wang, X.~Jiang, S.~Galliani, C.~Vogel, and M.~Pollefeys, ``{GLACE}: Global
  local accelerated coordinate encoding,'' in \emph{Proc. IEEE/CVF Conf.
  Comput. Vis. Pattern Recognit. (CVPR)}, 2024, pp. 21\,562--21\,571.

\bibitem{jiang2025rscore}
X.~Jiang, F.~Wang, S.~Galliani, C.~Vogel, and M.~Pollefeys, ``{R-SCoRe}:
  Revisiting scene coordinate regression for robust large-scale visual
  localization,'' in \emph{Proc. IEEE/CVF Conf. Comput. Vis. Pattern Recognit.
  (CVPR)}, 2025, pp. 11\,536--11\,546.

\bibitem{jang2017gumbel}
E.~Jang, S.~Gu, and B.~Poole, ``Categorical reparameterization with
  {Gumbel-Softmax},'' in \emph{Proc. Int. Conf. Learn. Represent. (ICLR)},
  2017.

\bibitem{wang2024auxfree}
L.~Wang, H.~Gao, C.~Zhao, X.~Sun, and D.~Dai, ``Auxiliary-loss-free load
  balancing strategy for mixture-of-experts,'' arXiv:2408.15664, 2024.

\bibitem{riquelme2021vmoe}
C.~Riquelme, J.~Puigcerver, B.~Mustafa, M.~Neumann, R.~Jenatton,
  A.~Susano~Pinto, D.~Keysers, and N.~Houlsby, ``Scaling vision with sparse
  mixture of experts,'' in \emph{Proc. Adv. Neural Inf. Process. Syst.
  (NeurIPS)}, 2021, pp. 8583--8595.

\bibitem{fischler1981ransac}
M.~A. Fischler and R.~C. Bolles, ``Random sample consensus: A paradigm for
  model fitting with applications to image analysis and automated
  cartography,'' \emph{Commun. ACM}, vol.~24, no.~6, pp. 381--395, 1981.

\bibitem{chum2003lo}
O.~Chum, J.~Matas, and J.~Kittler, ``Locally optimized {RANSAC},'' in
  \emph{Proc. DAGM Symp. Pattern Recognit.}, 2003, pp. 236--243.

\bibitem{barath2020magsacpp}
D.~Barath, J.~Noskova, M.~Ivashechkin, and J.~Matas, ``{MAGSAC++}, a fast,
  reliable and accurate robust estimator,'' in \emph{Proc. IEEE/CVF Conf.
  Comput. Vis. Pattern Recognit. (CVPR)}, 2020, pp. 1304--1312.

\bibitem{yang2020gnc}
H.~Yang, P.~Antonante, V.~Tzoumas, and L.~Carlone, ``Graduated non-convexity
  for robust spatial perception: From non-minimal solvers to global outlier
  rejection,'' \emph{IEEE Robot. Autom. Lett.}, vol.~5, no.~2, pp. 1127--1134,
  2020.

\bibitem{woo2018cbam}
S.~Woo, J.~Park, J.-Y. Lee, and I.~S. Kweon, ``{CBAM}: Convolutional block
  attention module,'' in \emph{Proc. Eur. Conf. Comput. Vis. (ECCV)}, 2018, pp.
  3--19.

\bibitem{vaswani2017attention}
A.~Vaswani, N.~Shazeer, N.~Parmar, J.~Uszkoreit, L.~Jones, A.~N. Gomez,
  {\L}.~Kaiser, and I.~Polosukhin, ``Attention is all you need,'' in
  \emph{Proc. Adv. Neural Inf. Process. Syst. (NeurIPS)}, 2017, pp. 5998--6008.

\bibitem{wang2018nonlocal}
X.~Wang, R.~Girshick, A.~Gupta, and K.~He, ``Non-local neural networks,'' in
  \emph{Proc. IEEE/CVF Conf. Comput. Vis. Pattern Recognit. (CVPR)}, 2018, pp.
  7794--7803.

\bibitem{touvron2021cait}
H.~Touvron, M.~Cord, A.~Sablayrolles, G.~Synnaeve, and H.~J{\'e}gou, ``Going
  deeper with image transformers,'' in \emph{Proc. IEEE/CVF Int. Conf. Comput.
  Vis. (ICCV)}, 2021, pp. 32--42.

\bibitem{huang2016droppath}
G.~Huang, Y.~Sun, Z.~Liu, D.~Sedra, and K.~Q. Weinberger, ``Deep networks with
  stochastic depth,'' in \emph{Proc. Eur. Conf. Comput. Vis. (ECCV)}, 2016, pp.
  646--661.

\bibitem{chen2022aspanformer}
H.~Chen, Z.~Luo, L.~Zhou, Y.~Tian, M.~Zhen, T.~Fang, D.~McKinnon, Y.~Tsin, and
  L.~Quan, ``{ASpanFormer}: Detector-free image matching with adaptive span
  transformer,'' in \emph{Proc. Eur. Conf. Comput. Vis. (ECCV)}, 2022, pp.
  20--36.

\bibitem{marquardt1963lm}
D.~W. Marquardt, ``An algorithm for least-squares estimation of nonlinear
  parameters,'' \emph{J. Soc. Ind. Appl. Math.}, vol.~11, no.~2, pp. 431--441,
  1963.

\bibitem{brachmann2021limits}
E.~Brachmann, M.~Humenberger, C.~Rother, and T.~Sattler, ``On the limits of
  pseudo ground truth in visual camera re-localisation,'' in \emph{Proc.
  IEEE/CVF Int. Conf. Comput. Vis. (ICCV)}, 2021, pp. 6218--6228.

\bibitem{sarlin2022lamar}
P.-E. Sarlin, M.~Dusmanu, J.~L. Sch{\"o}nberger, P.~Speciale, L.~Gruber,
  V.~Larsson, O.~Miksik, and M.~Pollefeys, ``{LaMAR}: Benchmarking localization
  and mapping for augmented reality,'' in \emph{Proc. Eur. Conf. Comput. Vis.
  (ECCV)}, 2022, pp. 686--704.

\end{thebibliography}

\end{document}